# Evolution and Efficiency in Neural Architecture Search: Bridging the Gap between Expert Design and Automated Optimization

**Fanfei Meng\***, **Chen-Ao Wang and Lele Zhang**

*Northwestern University, United State*

**\*Corresponding Author**
Fanfei Meng, Northwestern University, United State.

**Abstract**
*The paper provides a comprehensive overview of Neural Architecture Search (NAS), emphasizing its evolution from manual design to automated, computationally driven approaches. It covers the inception and growth of NAS, highlighting its application across various domains, including medical imaging and natural language processing. The document details the shift from expert-driven design to algorithm-driven processes, exploring initial methodologies like reinforcement learning and evolutionary algorithms. It also discusses the challenges of computational demands and the emergence of efficient NAS methodologies, such as Differentiable Architecture Search and hardware-aware NAS. The paper further elaborates on NAS's application in computer vision, NLP, and beyond, demonstrating its versatility and potential for optimizing neural network architectures across different tasks. Future directions and challenges, including computational efficiency and the integration with emerging AI domains, are addressed, showcasing NAS's dynamic nature and its continued evolution towards more sophisticated and efficient architecture search methods.*

**Keywords:** Neural Architecture Search, Automated Design Optimization, Computational Efficiency, Evolutionary Algorithms, Differentiable Architecture Search.

## 1. Introduction

The inception and early development of Neural Architecture Search (NAS) represent a transformative phase in artificial intelligence, particularly in deep learning. The quest for automating the design of neural network architectures has seen significant milestones, with research efforts focusing on overcoming the limitations of manual architecture design and leveraging computational strategies to discover optimal network structures. Early research in the domain of NAS was marked by efforts to understand and improve recurrent neural networks, such as the Long Short-Term Memory (LSTM) networks [1]. Conducted one of the largest studies on LSTM variants, assessing their utility and optimizing hyperparameters through random search, underscoring the early interest in automating aspects of neural network design.

Similarly, introduced the Inception architecture, demonstrating the potential of carefully crafted design to utilize computing resources more efficiently, a principle that would later influence NAS approaches [2]. The growing interest in NAS also extended to the medical imaging field, where the capabilities of deep learning models, including those designed through NAS, showed potential for surpassing human performance in certain recognition tasks, as highlighted by [3,4]. Further exemplified the evolution of NAS into specialized domains such as natural language processing, introducing NAS-Bench-NLP to facilitate research in the area by providing a comprehensive benchmark.

The trajectory of NAS research from its early days to its current status underscores a broad and ambitious effort to automate and optimize the design of neural networks across various domains. From enhancing LSTM networks to pioneering in the convolutional neural network (CNN) architectures and extending to medical and language processing applications, NAS embodies the transition from manual, expert-driven design to automated, computationally-driven architecture search processes. This evolution marks a significant shift towards democratizing and accelerating innovation in deep learning, promising to unlock new capabilities and efficiencies in AI systems.

### 1.1 Shift from Expert-Driven to Automated Design
Initially, neural network design was predominantly expert-driven, requiring a deep understanding of how different architectural choices, like the number of layers, types of layers (e.g., convolutional, recurrent), and connections between layers, affect the network's performance. This expertise was hard-earned through extensive experimentation. NAS shifted this paradigm by introducing algorithms capable of exploring vast architectural spaces, identifying optimal configurations often overlooked in manual processes. The first models and algorithms in NAS were foundational, setting the stage for more sophisticated techniques. These early efforts explored simple strategies like grid search and random search, quickly evolving to more complex approaches like reinforcement learning, evolutionary algorithms, and gradient-based methods. The evolution of NAS reflects a broader trend in AI: the move towards systems that can learn and adapt autonomously, reducing reliance on human input. NAS sits at the intersection of several disciplines: machine learning, optimization, statistics, and computational theory. Its development has been bolstered by advances in each of these areas, benefitting from improved computational resources, theoretical understanding of deep learning models, and the growing availability of large datasets. This multidisciplinary nature has made NAS a vibrant and rapidly advancing field, attracting attention from both academia and industry.

## 2. Backgrounds and Related Work
### 2.1 Reducing Human Effort in Network Design
The fundamental aim of Neural Architecture Search (NAS) is to mitigate the significant manual effort and expertise historically required in the development of neural network architectures. Traditionally, the design of effective neural networks has been a domain of specialists, necessitating a prolonged process of trial and error to identify optimal configurations. NAS introduces a revolutionary methodology with the goal of automating and refining this design process. By leveraging advanced algorithms, NAS endeavors to democratize neural network design, making it more universally accessible and diminishing the dependency on specialized knowledge. This approach not only accelerates the discovery of efficient neural architectures but also broadens the scope of innovation by exploring a vast array of design possibilities that might be beyond human intuition.

2.2 Exploring a Larger Space of Architectural Possibilities
The traditional approach to designing neural networks has been significantly limited by its reliance on human expertise, which inherently restricts the exploration of the vast architectural space to the realms of prior experience and intuitive guesswork. This method often misses out on discovering innovative and more efficient architectures that lie beyond conventional wisdom. Neural Architecture Search (NAS) emerged as a revolutionary solution to this issue, harnessing powerful algorithms to systematically explore a much wider range of architectural possibilities. By doing so, NAS opens the door to identifying novel neural network configurations that, while possibly counterintuitive to human designers, could offer vastly superior performance and efficiency. This comprehensive exploration facilitated by NAS not only broadens the horizon of potential designs but also democratizes the design process, making the development of high-performing neural networks more accessible and less constrained by the bottleneck of specialized knowledge.

### 2.3 Initial Methods and Computational Challenges
The early methods in Neural Architecture Search (NAS) utilized strategies such as Reinforcement Learning (RL) and Evolutionary Algorithms (EAs) to explore and optimize neural network architectures. RL-based NAS approaches employed policy networks to generate and refine architectures based on performance feedback, whereas EAs drew inspiration from biological evolution, iterating over generations of architectures to enhance performance through selection, mutation, and recombination. Despite their innovativeness, these initial strategies demanded extensive computational resources, which limited their practical application. Evolutionary design methods, particularly for deep convolutional neural networks (CNNs) in image classification, underscored the importance of multi-objective optimization that balances classification performance with computational costs[4]. Subsequent research has introduced more efficient NAS methodologies to address these computational challenges [5].

Adaptive Scalable NAS (AS-NAS): This method combines a simplified RL algorithm with the reinforced I-Ching divination evolutionary algorithm (IDEA) and a variable-architecture encoding strategy for efficient operator selection and scalable deep neural architecture design. The integration with L2 regularization enhances architecture sparsity, significantly reducing computational costs while maintaining or improving performance [6]. Performance Predictors for ENAS: Introduces a novel training protocol for performance predictors in evolutionary NAS, addressing the high computational demand by improving prediction accuracy of architecture performance without extensive computational resources. This method employs a pairwise ranking indicator, logistic regression for fitting training samples, and a differential method for constructing training instances, significantly enhancing the efficiency of NAS [7]. Hill Climbing-Based NAS Framework: Proposes a new framework using hill-climbing procedures and morphism operators, reducing overall training time by focusing on the aging of neural network layers. This approach demonstrates competitive results with significantly less computational expense [8]. Efficient Evolutionary Search of Attention Convolutional Networks: This framework introduces a computationally efficient evolutionary search method for convolutional networks, incorporating sampled training and node inheritance to evaluate offspring individuals without extensive training. The inclusion of a channel attention mechanism in the search space further enhances feature processing capability [9].

### 2.4 Overcoming Computational Limitations
To tackle the computational challenges inherent in Neural Architecture Search (NAS), researchers have pursued more efficient methodologies, highlighting innovations beyond early adaptations like DARTS (Differentiable Architecture Search). The DARTS+ algorithm exemplifies this evolution by refining DARTS through the incorporation of early stopping, which curtails the optimization process to prevent overfitting—a common pitfall that undermines the model's generalization ability. This modification significantly enhances the robustness of the architecture search process, avoiding the detrimental performance collapse frequently observed in conventional DARTS implementations [10].

Moreover, alternative strategies have emerged to surpass the limitations of early-stopping mechanisms. Notably, the approach of Random Search with weight-sharing has demonstrated remarkable effectiveness, challenging the prevailing complexity of NAS methodologies. By leveraging a simpler, more stochastic process, this method has not only matched but, in some instances, outperformed the sophisticated algorithms typically associated with NAS, showcasing its prowess on benchmarks such as PTB and CIFAR-10. The success of Random Search with weight-sharing underscores the potential of resource-efficient methods to achieve competitive, if not superior, NAS outcomes, thereby reshaping the landscape of architectural search paradigms [11]. The exploration of NAS methodologies extends into diverse adaptations and refinements aimed at overcoming the computational hurdles initially posed by traditional approaches. For instance, Fair DARTS introduces a collaborative rather than exclusive competition among operations by allowing each operation's architectural weight to be independent. This strategy effectively mitigates the performance collapse associated with skip connections, a common issue in DARTS, thereby promoting a more equitable and balanced search process that yields state-of-the-art results on CIFAR-10 and ImageNet [12]. D-DARTS, on the other hand, expands the search space by nesting neural networks at the cell level, diverging from the conventional weight-sharing mechanism. This innovation allows for the production of diversified and specialized architectures, highlighting the potential to enhance performance and reduce computational time across various computer vision tasks [13].

These advancements represent a pivotal shift towards more efficient, effective, and equitable NAS methodologies. By addressing the inherent challenges of computational intensity and overfitting, the field moves closer to realizing the full potential of automated architecture search in advancing deep learning innovations.

### 3. Architecture Search Mechanism
#### 3.1 Defining the Search Space
The foundation of Neural Architecture Search (NAS) lies in accurately defining the search space, which is a comprehensive catalog of all potential neural network architectures that the algorithm might evaluate. This space is multi-dimensional and encompasses a wide array of architectural choices, such as the types of layers (e.g., convolutional layers for image tasks, recurrent layers for sequential data), the connectivity patterns between these layers (e.g., sequential, skip connections), hyperparameters (including the depth of the network, sizes of filters in convolutional layers, and number of units in dense layers), and other architectural features like activation functions and regularization techniques. The breadth and depth of the search space are critical because they directly influence the NAS algorithm's ability to discover innovative and high-performing architectures. A well-constructed search space should balance comprehensiveness with feasibility, ensuring a wide range of architectures can be explored without making the search impractical due to computational constraints.

#### 3.2 Choosing a Search Strategy
The strategy for navigating the vast search space is pivotal in the NAS process. Various methodologies offer different trade-offs in terms of search efficiency, computational demands, and the quality of the resulting architectures. Reinforcement learning approaches, for example, use a policy network to sequentially choose architectural components, learning to propose better architectures based on past performance. Evolutionary algorithms simulate a process akin to natural selection, where architectures mutate and crossover, with only the fittest surviving to the next generation. Gradient-based methods allow for a more direct approach by optimizing architecture parameters using gradient descent, making them efficient but sometimes less exploratory. Bayesian optimization leverages prior knowledge to smartly explore the search space, balancing between exploitation of known good areas and exploration of new ones. The choice among these strategies depends on the specific goals, computational resources available, and the characteristics of the problem at hand.

#### 3.3 Performance Evaluation Mechanism
A crucial step in NAS is the evaluation of candidate architectures. This step determines how well each proposed architecture performs on a given task, which in turn informs the search algorithm's future decisions. Typically, the evaluation involves training the architecture on a dataset and measuring its performance using metrics such as accuracy, loss, or more task-specific measures. However, training neural networks from scratch for each evaluation can be prohibitively expensive. Techniques like weight sharing, where different architectures share weights, and proxy tasks, where models are trained and evaluated on simplified versions of the task or with reduced data, can significantly reduce computational requirements. This step not only identifies promising architectures but also helps in understanding the impact of different architectural choices on performance.

#### 3.4 Optimization Process
With the performance feedback from the evaluation mechanism, the NAS algorithm optimizes the search towards architectures that show potential for high performance. This optimization can take various forms depending on the search strategy. For instance, reinforcement learning algorithms adjust the policy network to increase the probability of selecting high-performing architectures. In contrast, evolutionary algorithms might adjust the population towards architectures with higher fitness scores. This optimization process is iterative, with each cycle aiming to refine the search direction and converge towards optimal architectural configurations. The ultimate goal is to discover architectures that not only perform well on the evaluation metric(s) but also meet other criteria such as efficiency in terms of computational resources and model size.

#### 3.5 Refinement and Final Selection
After identifying a subset of promising architectures through the search and optimization process, a refinement phase may follow. This phase involves further fine-tuning the architectures, possibly through additional rounds of training, hyperparameter optimization, or architectural tweaks, to squeeze out additional performance gains. The final selection of the optimal architecture is then made based on a comprehensive assessment of performance metrics, alongside other considerations like computational efficiency, model complexity, and adaptability to different tasks or datasets. This step ensures that the chosen architecture not only performs well in a controlled evaluation setting but is also practical for real-world applications.

### 3.6 Verification and Testing
The selected architecture undergoes rigorous testing to verify its performance, generalizability across different datasets or tasks, and robustness to variations in input or conditions. This step is essential to ensure that the model's performance is not overly specific to the training set or evaluation conditions used during the NAS process. Extensive testing provides confidence that the chosen architecture will perform reliably in practical applications, fulfilling the promise of NAS to automate the design of effective and efficient neural networks.

### 3.7 Deployment
Once verified, the final architecture is ready for deployment in real-world applications. This could range from image and speech recognition systems in consumer electronics to complex decision-making systems in autonomous vehicles or personalized recommendation systems in e-commerce. The deployment phase marks the culmination of the NAS process, translating the computational and algorithmic achievements into tangible benefits in various applications.

The iterative nature of NAS, with its cycles of exploration, evaluation, optimization, and refinement, encapsulates a comprehensive approach to automating neural network design. While computationally demanding, NAS holds the promise of discovering innovative architectures that push the boundaries of what is possible in artificial intelligence, offering efficiency and performance improvements over traditional, manually designed models.

### 4. Mainstreaming Paradigm of NAS
1. Reinforcement Learning-based NAS (RL-NAS): RL-NAS employs a policy network as the decision-maker to sequentially propose neural network architectures. This policy network, trained through reinforcement learning techniques such as Q-learning or policy gradients, learns to navigate the architecture search space by receiving feedback on the performance (e.g., accuracy, efficiency) of its suggestions. The iterative process of proposing architectures, evaluating their performance on a validation set, and updating the policy network based on this feedback allows RL-NAS to refine its search strategy over time. Despite its potential to discover high-performing architectures, RL-NAS is noted for its high computational demand, as each proposed architecture requires separate training and evaluation, making it resource-intensive.
2. Evolutionary Algorithms (EA): Evolutionary algorithms mimic the process of natural selection by applying operations such as mutation, crossover (recombination of architectural elements), and selection to evolve a population of neural network architectures over generations. Each architecture in the population is evaluated based on its performance, and the best-performing architectures are more likely to be selected for breeding the next generation. This process encourages the exploration of the architecture space and the discovery of innovative solutions. However, like RL-NAS, EA-based NAS is computationally expensive due to the extensive evaluations needed across generations of architectures.
3. Differentiable NAS (D-NAS): Differentiable NAS stands out by making the architecture search space continuous, which allows the use of gradient descent for optimization. Techniques like DARTS (Differentiable Architecture Search) introduce architectural parameters that can be optimized alongside the network weights, enabling simultaneous learning of the architecture and its parameters. This approach dramatically accelerates the search process compared to non-differentiable methods. D-NAS is celebrated for its efficiency and has been instrumental in reducing the computational barrier to performing NAS, making it accessible for more researchers and practitioners.
4. Bayesian Optimization-based NAS: Bayesian Optimization (BO) is a strategy for global optimization of black-box functions that is particularly effective in situations where function evaluations (architecture evaluations, in the context of NAS) are costly. BO-based NAS uses probabilistic models to predict the performance of architectures and applies an acquisition function to balance the exploration of new architectures with the exploitation of known good ones. This method efficiently navigates the search space by prioritizing the evaluation of architectures that are most likely to yield improvements, making it suitable for scenarios with limited computational resources.
5. Graph-based NAS: Utilizing graph theory, graph-based NAS represents the architecture search space as a graph, where nodes represent architectural components (e.g., layers, operations) and edges represent connections between these components. This representation allows for the modeling of complex architectures with varying depths and connectivity patterns. Graph-based methods can efficiently explore this space using algorithms that manipulate the graph structure, offering a versatile approach to discovering architectures that can capture intricate data patterns.
6. Hierarchical NAS: Hierarchical NAS addresses the architecture search at multiple levels of granularity, distinguishing between the macro-architecture (the overall structure and connectivity of the network) and the micro-architecture (the design of individual layers or blocks within the network). By optimizing both levels, hierarchical NAS enables a more detailed and nuanced exploration of the architecture space, potentially leading to more optimized and task-specific designs that consider both the global structure and the local operations of the network.
7. One-shot NAS: One-shot NAS methods streamline the search process by constructing a single, over-parameterized network (often called a supernet) that encompasses all possible architectures within the search space. By training this supernet and then evaluating sub-networks (candidate architectures) without retraining from scratch, one-shot NAS significantly reduces the computational cost of NAS. This approach benefits from weight sharing among sub-networks, allowing for rapid assessment of numerous architectures.
8. Cell-based NAS: Focusing on finding optimal building blocks (cells), which are then replicated to construct the entire network, cell-based NAS simplifies the search space and concentrates on discovering versatile and reusable architectural patterns. This strategy has proven effective in identifying architectures that perform well across different datasets and tasks, as the discovered cells can be adapted to various scales and complexities of problems.Neuroevolution: Neuroevolution combines the principles of evolutionary algorithms with neural networks to co-evolve both the architecture and the weights of the network. This approach treats the architecture and weights as part of a unified evolutionary process, exploring a wide range of potential solutions by mutating, recombining, and selecting networks based on their performance. Neuroevolution is particularly appealing for its ability to discover both innovative architectures and optimal weight configurations, offering a holistic approach to network design and optimization.

Each of these NAS methodologies offers unique advantages and trade-offs, reflecting the diversity of challenges and objectives present in neural network design. As NAS continues to evolve, the interplay among these approaches will likely yield even more sophisticated and efficient methods for automated architecture search.

## 5. Development and Elaboration

NAS has made significant advancements in recent years, addressing its initial shortcomings and introducing more sophisticated and efficient methods:
1. Hardware-Aware NAS: Recent approaches like hardware-aware frameworks have focused on optimizing NAS not only for model accuracy but also for hardware efficiency. These methods use evolutionary algorithms paired with objective predictors to efficiently find optimized architectures for various performance metrics and hardware configurations [14].
2. NAS-Bench-101 for Reproducibility: The introduction of NAS-Bench-101, the first public architecture dataset for NAS research, aims to address the high computational demand of NAS and make experiments more reproducible. It compiles a large dataset of over 5 million trained models, allowing rapid evaluation of a diverse range of architectures [15].
3. Optimization for Embedded Devices:** Adaptations of methods like Efficient Neural Architecture Search (ENAS) consider constraints for deploying networks on embedded devices, demonstrating NAS's flexibility and applicability in resource-constrained environments[16].
4. Pareto-Optimal Approaches: Multi-objective frameworks like MONAS and DPP-Net extend NAS to optimize for accuracy and other objectives imposed by devices, searching for neural architectures that can be deployed across a wide spectrum of devices [17].
5. Efficient NAS for Image Denoising:** Developments in NAS have also been applied to image denoising, with techniques like superkernel implementations enabling fast training of models for dense predictions, showcasing NAS's versatility in different application domains [18].
6. Evolutionary Design for Image Classification:** Evolutionary algorithms have been used to design deep convolutional neural networks for image classification, addressing multiple objectives such as classification performance and computational efficiency [19]. These advancements demonstrate the dynamic nature of NAS, continually evolving to address computational efficiency, hardware constraints, and application-specific requirements. The field is moving towards more versatile, efficient, and application-tailored architecture search methods, paving the way for broader adoption and more innovative applications of NAS.

## 6. Application of NAS

In the realm of computer vision and beyond [20-22], Neural Architecture Search (NAS) has emerged as a transformative force, propelling advancements in image classification, object detection, semantic segmentation, and Natural Language Processing (NLP). The integration of frameworks like DQNAS, which combines reinforcement learning with one-shot training, illustrates NAS's capacity to surpass manually designed models across various image-related applications. This approach not only streamlines the creation of efficient models for complex tasks such as language modeling and translation but also extends its utility to optimizing convolutional networks for feature extraction at multiple resolutions [23-26]. The adaptability of NAS is further demonstrated in its application to Human Activity Recognition (HAR), where it refines neural architectures for analyzing mobility-related human activities using techniques like Bayesian optimization. This showcases NAS's effectiveness in specialized domains, underscoring its role in enhancing the accuracy and efficiency of models tailored for specific tasks and data types [27].

Expanding its influence, NAS plays a pivotal role in spatio-temporal prediction tasks within smart city applications, exemplified by methods like AutoST for crowd flow prediction, highlighting NAS's contribution to urban planning and intelligent transportation. This broad applicability of NAS across different fields—from enhancing machine interpretation of visual information to improving interaction with human language, and advancing our understanding of human movement through technology—emphasizes its integral role in automating the design of neural network architectures. As NAS continues to evolve [29-30], it promises to further drive innovations and optimizations in neural network architectures, cementing its impact across diverse areas of AI and machine learning. The ongoing advancements underscore the potential of NAS to revolutionize various aspects of research and application, making it a cornerstone of future developments in technology [31-42].

## References
1. Greff, K., Srivastava, R. K., Koutník, J., Steunebrink, B. R., & Schmidhuber, J. (2016). LSTM: A search space odyssey. *IEEE transactions on neural networks and learning systems, 28*(10), 2222-2232.
2. Szegedy, C., Liu, W., Jia, Y., Sermanet, P., Reed, S., Anguelov, D., ... & Rabinovich, A. (2015). Going deeper with convolutions. In *Proceedings of the IEEE conference on computer vision and pattern recognition* (pp. 1-9).
3. Lee, J. G., Jun, S., Cho, Y. W., Lee, H., Kim, G. B., Seo, J. B., & Kim, N. (2017). Deep learning in medical imaging: general overview. *Korean journal of radiology, 18*(4), 570-584.
4. Klyuchnikov, N., Trofimov, I., Artemova, E., Salnikov, M., Fedorov, M., Filippov, A., & Burnaev, E. (2022). Nas-

bench-nlp: neural architecture search benchmark for natural language processing. *IEEE Access, 10*, 45736-45747.
5. Lu, Z., Whalen, I., Dhebar, Y., Deb, K., Goodman, E. D., Banzhaf, W., & Boddeti, V. N. (2020). Multiobjective evolutionary design of deep convolutional neural networks for image classification. *IEEE Transactions on Evolutionary Computation, 25*(2), 277-291.
6. Zhang, T., Lei, C., Zhang, Z., Meng, X. B., & Chen, C. P. (2021). AS-NAS: Adaptive scalable neural architecture search with reinforced evolutionary algorithm for deep learning. *IEEE Transactions on Evolutionary Computation, 25*(5), 830-841.
7. Sun, Y., Sun, X., Fang, Y., Yen, G. G., & Liu, Y. (2021). A novel training protocol for performance predictors of evolutionary neural architecture search algorithms. *IEEE Transactions on Evolutionary Computation, 25*(3), 524-536.
8. Verma, M., Sinha, P., Goyal, K., Verma, A., & Susan, S. (2019, June). A novel framework for neural architecture search in the hill climbing domain. In *2019 IEEE Second International Conference on Artificial Intelligence and Knowledge Engineering (AIKE)* (pp. 1-8). IEEE.
9. Zhang, H., Jin, Y., Cheng, R., & Hao, K. (2020). Efficient evolutionary search of attention convolutional networks via sampled training and node inheritance. *IEEE Transactions on Evolutionary Computation, 25*(2), 371-385.
10. Liang, H., Zhang, S., Sun, J., He, X., Huang, W., Zhuang, K., & Li, Z. (2019). Darts+: Improved differentiable architecture search with early stopping. *arXiv preprint arXiv:1909.06035*.
11. Li, L., & Talwalkar, A. (2020, August). Random search and reproducibility for neural architecture search. In *Uncertainty in artificial intelligence* (pp. 367-377). PMLR.
12. Chu, X., Zhou, T., Zhang, B., & Li, J. (2020, August). Fair darts: Eliminating unfair advantages in differentiable architecture search. In *European conference on computer vision* (pp. 465-480). Cham: Springer International Publishing.
13. Heuillet, A., Tabia, H., Arioui, H., & Youcef-Toumi, K. (2023). D-DARTS: Distributed differentiable architecture search. *Pattern Recognition Letters, 176*, 42-48.
14. Cummings, D., Sarah, A., Sridhar, S. N., Szankin, M., Munoz, J. P., & Sundaresan, S. (2022). A hardware-aware framework for accelerating neural architecture search across modalities. *arXiv preprint arXiv:2205.10358*.
15. Ying, C., Klein, A., Christiansen, E., Real, E., Murphy, K., & Hutter, F. (2019, May). Nas-bench-101: Towards reproducible neural architecture search. In *International conference on machine learning* (pp. 7105-7114). PMLR.
16. Cassimon, T., Vanneste, S., Bosmans, S., Mercelis, S., & Hellinckx, P. (2020). Using neural architecture search to optimize neural networks for embedded devices. In *Advances on P2P, Parallel, Grid, Cloud and Internet Computing: Proceedings of the 14th International Conference on P2P, Parallel, Grid, Cloud and Internet Computing (3PGCIC-2019) 14* (pp. 684-693). Springer International Publishing.
17. Cheng, A. C., Dong, J. D., Hsu, C. H., Chang, S. H., Sun, M., Chang, S. C., ... & Juan, D. C. (2018, November). Searching toward pareto-optimal device-aware neural architectures. In *Proceedings of the international conference on computer-aided design* (pp. 1-7).
18. Mozejko, M., Latkowski, T., Treszczotko, L., Szafraniuk, M., & Trojanowski, K. (2020). Superkernel neural architecture search for image denoising. In *Proceedings of the IEEE/CVF Conference on Computer Vision and Pattern Recognition Workshops* (pp. 484-485).
19. Lu, Z., Whalen, I., Dhebar, Y., Deb, K., Goodman, E. D., Banzhaf, W., & Boddeti, V. N. (2020). Multiobjective evolutionary design of deep convolutional neural networks for image classification. *IEEE Transactions on Evolutionary Computation, 25*(2), 277-291.
20. Meng, F., & Wang, Y. (2023). Transformers: Statistical interpretation, architectures and applications. *Authorea Preprints*.
21. Meng, B. G. F., & Ghena, B. (2023). Research on text recognition methods based on artificial intelligence and machine learning. *preprint under review*.
22. Meng, F., & Demeter, D. (2023). Sentiment analysis with adaptive multi-head attention in Transformer. *arXiv preprint arXiv:2310.14505*.
23. Razeghi, M., Dehzangi, A., Wu, D., McClintock, R., Zhang, Y., Durlin, Q., ... & Meng, F. (2019, May). Antimonite-based gap-engineered type-II superlattice materials grown by MBE and MOCVD for the third generation of infrared imagers. In *Infrared Technology and Applications XLV* (Vol. 11002, pp. 108-125). SPIE.
24. Meng, F., Zhang, L., Chen, Y., & Wang, Y. (2023). FedEmb: A Vertical and Hybrid Federated Learning Algorithm using Network And Feature Embedding Aggregation. *arXiv preprint arXiv:2312.00102*.
25. Meng, F., Zhang, L., Chen, Y., & Wang, Y. (2023). Sample-based Dynamic Hierarchical Transformer with Layer and Head Flexibility via Contextual Bandit. *Authorea Preprints*.
26. Meng, F., & Wang, C. A. (2023). A Dynamic Interactive Learning Interface for Computer Science Education: Programming Decomposition Tool. *Authorea Preprints*.
27. Ling, C., Zhang, C., Wang, M., Meng, F., Du, L., & Yuan, X. (2020). Fast structured illumination microscopy via deep learning. *Photonics Research, 8*(8), 1350-1359.
28. Meng, F., Jagadeesan, L., & Thottan, M. (2021). Model-based reinforcement learning for service mesh fault resiliency in a web application-level. *arXiv preprint arXiv:2110.13621*.
29. Wang, Y., Meng, F., Wang, X., & Xie, C. (2023). Optimizing the Passenger Flow for Airport Security Check. *arXiv preprint arXiv:2312.05259*.
30. Chen, J. J., Xu, Q., Wang, T., Meng, F. F., Li, Z. W., Fang, L. Q., & Liu, W. (2022). A dataset of diversity and distribution of rodents and shrews in China. *Scientific Data, 9*(1), 304.
31. Meng, F., Zhang, L., Wang, Y., & Zhao, Y. (2023). Joint detection algorithm for multiple cognitive users in spectrum sensing. *Authorea Preprints*.
32. Meng, B. G. F., & Ghena, B. (2023). Research on text recognition methods based on artificial intelligence and machine learning. *preprint under review*.

33. Meng, F., & Demeter, D. (2023). Sentiment analysis with adaptive multi-head attention in Transformer. *arXiv preprint arXiv:2310.14505.*
34. Razeghi, M., Dehzangi, A., Wu, D., McClintock, R., Zhang, Y., Durlin, Q., ... & Meng, F. (2019, May). Antimonite-based gap-engineered type-II superlattice materials grown by MBE and MOCVD for the third generation of infrared imagers. In *Infrared Technology and Applications XLV* (Vol. 11002, pp. 108-125). SPIE.
35. Meng, F., Zhang, L., Chen, Y., & Wang, Y. (2023). FedEmb: A Vertical and Hybrid Federated Learning Algorithm using Network And Feature Embedding Aggregation. *arXiv preprint arXiv:2312.00102.*
36. Meng, F., Zhang, L., Chen, Y., & Wang, Y. (2023). Sample-based Dynamic Hierarchical Transformer with Layer and Head Flexibility via Contextual Bandit. *Authorea Preprints.*
37. Meng, F., & Wang, C. A. (2023). A Dynamic Interactive Learning Interface for Computer Science Education: Programming Decomposition Tool. *Authorea Preprints.*
38. Ling, C., Zhang, C., Wang, M., Meng, F., Du, L., & Yuan, X. (2020). Fast structured illumination microscopy via deep learning. *Photonics Research, 8*(8), 1350-1359.
39. Meng, F., Jagadeesan, L., & Thottan, M. (2021). Model-based reinforcement learning for service mesh fault resiliency in a web application-level. *arXiv preprint arXiv:2110.13621.*
40. Wang, Y., Meng, F., Wang, X., & Xie, C. (2023). Optimizing the Passenger Flow for Airport Security Check. *arXiv preprint arXiv:2312.05259.*
41. Chen, J. J., Xu, Q., Wang, T., Meng, F. F., Li, Z. W., Fang, L. Q., & Liu, W. (2022). A dataset of diversity and distribution of rodents and shrews in China. *Scientific Data, 9*(1), 304.
42. Meng, F., Zhang, L., Wang, Y., & Zhao, Y. (2023). Joint detection algorithm for multiple cognitive users in spectrum sensing. *Authorea Preprints.*
43. Meng, F., & Wang, Y. (2023). Transformers: Statistical interpretation, architectures and applications. *Authorea Preprints.*